\documentclass[final]{cvpr}

\usepackage{times}
\usepackage{epsfig}
\usepackage{graphicx}
\usepackage{amsmath}
\usepackage{amssymb}
\usepackage{soul}

\usepackage[pagebackref=true,breaklinks=true,colorlinks,bookmarks=false]{hyperref}

\def\cvprPaperID{843} 
\def\confYear{CVPR 2021}

\usepackage[export]{adjustbox}
\usepackage[dvipsnames]{xcolor}
\usepackage{multirow}

\definecolor{purple}{RGB}{230,0,250}
\newcommand{\LXX}[1]{\textcolor{black}{{#1}}}

\usepackage{amssymb}
\usepackage{pifont}
\newcommand{\cmark}{\ding{51}}%
\newcommand{\xmark}{\ding{55}}%
\usepackage{environ}
\newcommand{\acksection}{\section*{Acknowledgments}}
\NewEnviron{acks}{%
  \acksection
  \BODY
}

\begin{document}

\title{Multi-view Depth Estimation using Epipolar Spatio-Temporal Networks}

\author{Xiaoxiao Long$^{1}$ \quad Lingjie Liu$^{2}$ \quad Wei Li$^{3}$ \quad Christian Theobalt$^{2}$ \quad Wenping Wang$^{1,4}$ \\[0.3em]
$^{1}$The University of Hong Kong \quad $^{2}$Max Planck Institute for Informatics \\
$^{3}$Inceptio \quad $^{4}$Texas A\&M University }


\maketitle
\thispagestyle{empty}

\begin{abstract}
We present a novel method for multi-view depth estimation from a single video, which is a critical task in various applications, such as perception, reconstruction and robot navigation. Although previous learning-based methods have demonstrated compelling results, most works estimate depth maps of individual video frames independently, without taking into consideration the strong geometric and temporal coherence among the frames. Moreover, current state-of-the-art (SOTA) models mostly adopt a fully 3D convolution network for cost regularization and therefore require high computational cost, thus limiting their deployment in real-world applications. Our method achieves temporally coherent depth estimation results by using a novel Epipolar Spatio-Temporal (EST) transformer to explicitly associate geometric and temporal correlation with multiple estimated depth maps. Furthermore, to reduce the computational cost, inspired by recent Mixture-of-Experts models, we design a compact hybrid network consisting of a 2D context-aware network and a 3D matching network which learn 2D context information and 3D disparity cues separately. Extensive experiments demonstrate that our method achieves higher accuracy in depth estimation and significant speedup than the SOTA methods. Code is available: \url{https://github.com/xxlong0/ESTDepth}.

\end{abstract}

\section{Introduction}

\begin{figure}[t]
\setlength{\abovecaptionskip}{0pt}
\setlength{\belowcaptionskip}{0pt}
    \centering
    \includegraphics[width=\columnwidth]{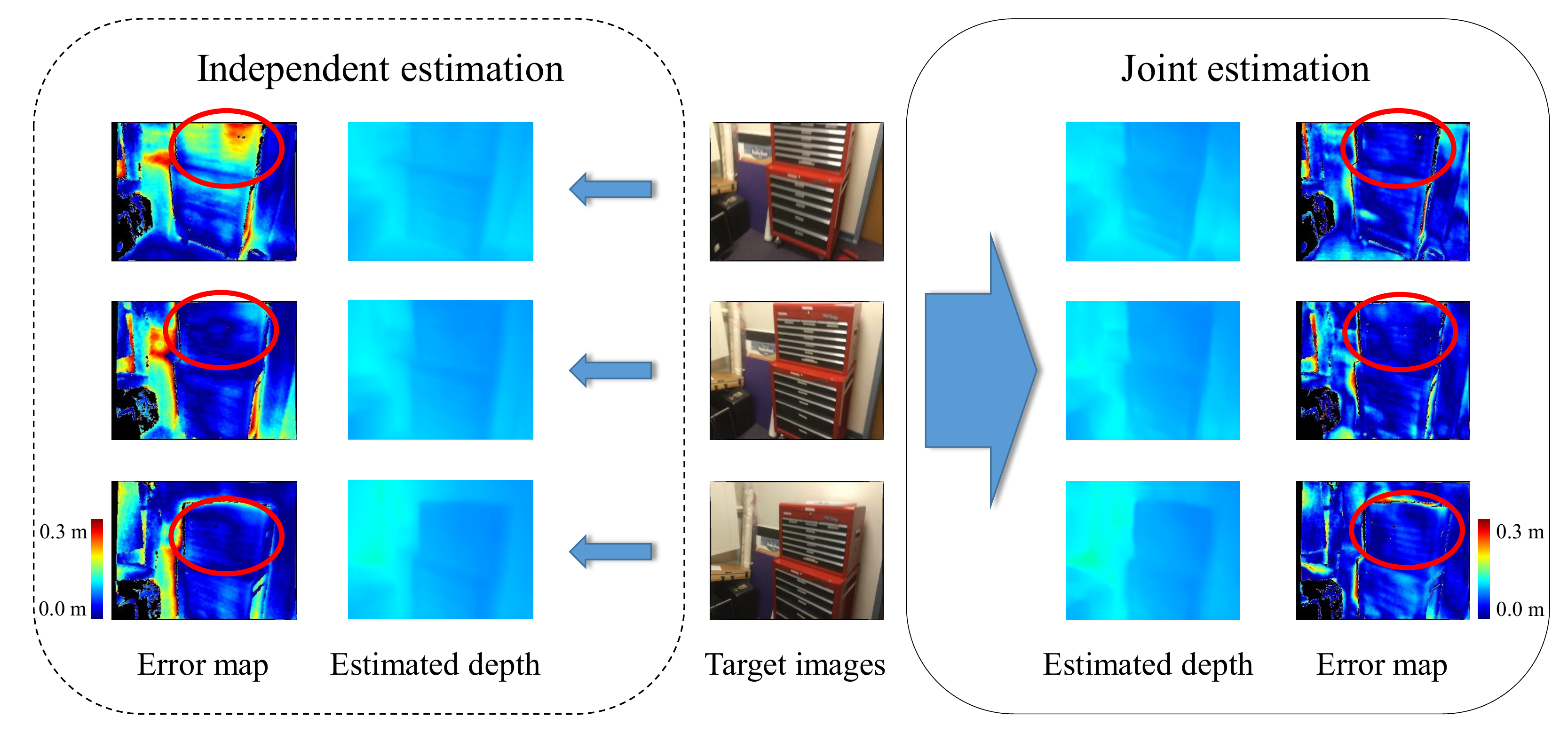}
    \caption{Independent estimation vs Joint estimation. Prior works 
 estimate the depth maps of frames from a video independently despite their temporal coherence, thus the estimated depth maps may contain inconsistent values for the same region (\textcolor{red}{red circles}). Our model could estimate multiple depth maps jointly utilizing temporal coherence, enabling the estimated depth maps to be consistent.}
    \label{fig:teaser}
\end{figure}

Multi-view depth estimation aims to recover 3D geometry of given images with known camera parameters, which is one of fundamental problems in computer vision. Many applications benefit from the recovered depth, such as autonomous driving~\cite{wang2019pseudo}, augmented reality~\cite{jones2008effects}, 3D modeling~\cite{henry2014rgb}, and image-based rendering~\cite{fehn2004depth}. 

Recently, learning based depth estimation methods, whether designed for images~\cite{eigen2014depth,liu2015deep,im2019dpsnet,yao2018mvsnet} or videos~\cite{liu2019neural,zhang2019exploiting,wang2019recurrent,patil2020don,luo2020consistent}, have achieved great improvements against their traditional counterparts~\cite{zhang2009consistent,schonberger2016structure,bleyer2011patchmatch,hosni2009local};
however, these methods, especially for videos, have significant room for improvement in terms of temporal coherence and computational efficiency. 

\textbf{Temporal coherence.} 
Most of multi-view depth estimation methods~\cite{im2019dpsnet,yao2018mvsnet,kusupati2020normal,long2020occlusion,wang2018mvdepthnet} are designed for individual images so they are not suitable for temporally coherent videos. Directly extending the existing methods from images to video sequences causes flickering artifacts. i.e. inconsistent estimated depth maps across consecutive frames, because they do not take temporal coherence into consideration. As shown in Figure.~\ref{fig:teaser}, the independently estimated depth maps contain inconsistent values. Therefore, it is necessary to jointly estimate depth maps of a video sequence to produce temporally coherent results.

Existing works on video depth estimation can be divided into two categories: single-view methods~\cite{zhang2019exploiting,cs2018depthnet,wang2019recurrent,patil2020don,luo2020consistent} and multi-view methods~\cite{liu2019neural}, according to the input of the depth estimation network. Recurrent neural units are widely used in single-view video methods~\cite{zhang2019exploiting,cs2018depthnet,wang2019recurrent,patil2020don} to encode temporal coherence implicitly. Although the inconsistency problem is alleviated by incorporating temporal coherence, these methods still suffer from the ambiguity of depth scale since depth estimation from a single image is an ill-posed problem. 

Multi-view video methods are advantageous over single-view methods because epipolar geometry information provided by multiple images could be used to avoid depth scale ambiguity problem.
To the best of our knowledge, there has been only one multi-view method~\cite{liu2019neural} for video depth estimation based on the epipolar geometry information.  This method produces compelling results but it is restricted by its specific network design that it can only 
\LXX{make use of one preceding estimation to improve current estimation.} 
We improve upon this multi-view method by proposing a novel Epipolar Spatio-Temporal (EST) network that is capable of utilizing temporal coherence of multiple preceding estimations, thus producing more accurate depth maps with better temporal coherence. 


\textbf{Computational Efficiency.}
Top-performing multi-view depth estimation models~\cite{im2019dpsnet,kusupati2020normal,liu2019neural} are slow with low computational efficiency, because they adopt a single fully 3D convolution networks for cost regularization to learn both local feature matching information and global context information, which have been shown to be critical for accurate depth estimation~\cite{chang2018pyramid, kendall2017end}. While the local feature information is necessary for matching texture-rich regions, the global context information is crucial for scenes with texture-less regions. 
The existing networks tend to use deeper and deeper networks to improve the ability of learning global context information, leading to increased computation cost. It should be pointed out that the global context information is essentially 2D, so learning it by deep 3D convolution layers will unnecessarily consume masses of computational resources.


Our insight is that for depth estimation the local feature information and global semantic information can be learned by two separate networks, as inspired by recent Mixture-of-Experts models~\cite{wang2020deep,zhang2019learning,ma2018modeling}. Specifically, we propose a hybrid cost regularization network, consisting of two complementary expert sub-networks: a 2D \textit{ContextNet} focusing on 2D global context information, and a shallow 3D \textit{MatchNet} concentrating on 3D local matching information. 
By explicitly disentangles these two different types of information, our hybrid network consumes much less GPU computational resources and achieves faster running speed.



Our main contributions are summarized as follows:
\begin{itemize}
  \item We proposed an Epipolar Spatio-Temporal (EST) transformer that propagates temporal coherence to perform joint depth estimation of multiple frames to make estimated depth maps more temporally coherent. 
  \item We designed a hybrid network for cost regularization that consists of two expert networks to learn 3D local matching information and 2D global context information separately. This decoupling approach achieves faster speed and consumes less computational resources than using a single fully 3D network in prior works.
  \item Based on these two contributions we developed a new mutli-view method for generating temporally coherent depth maps from videos. We conducted extensive experiments on several datasets to demonstrate that our method outperforms the SOTAs by a large margin in terms of accuracy and speed. 
\end{itemize}

\begin{figure*}[t]
    \centering
    \includegraphics[width=\textwidth]{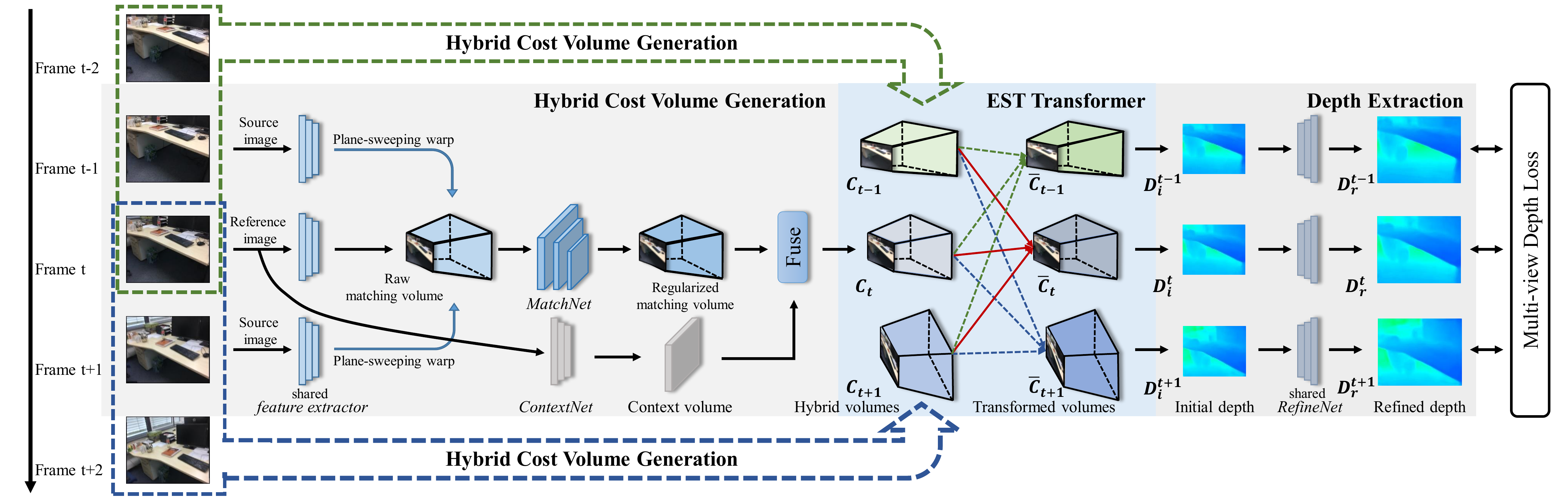}
    \caption{Overview of our method. In the training stage, our method takes a clip of video with five frames as input to estimate the depth maps of the middle three target frames. For each target image with its source images, we first extract learned feature maps by a shallow feature extractor. Subsequent plane-sweeping warping is applied to the feature maps to construct raw matching volume, and it is further regularized by \textit{MatchNet} to obtain a regularized matching volume. In parallel, we utilize \textit{ContextNet} to learn 2D context information from the target image, and concatenate context volume and matching volume together to get one hybrid volume. The Epipolar Spatio-Temporal (EST) transformer is applied on all obtained hybrid volumes to associate temporal coherence with them. Finally, we extract initial depth maps from the transformed volumes and further refine the initial depth maps by \textit{RefineNet}. }
    \label{fig:overview}
\end{figure*}

\section{Related Work}
\paragraph{Depth estimation from single image} Learning based single-view depth estimation methods~\cite{eigen2014depth,eigen2015predicting,chen2018rethinking,chen2016single,wang2015towards,xu2019depth,cheng2019learning,liu2020fcfr,liu2021learning} have been extensively studied in recent years. 
Convolution neural network for depth estimation is first introduced by \cite{eigen2014depth}, which demonstrates the superior ability of CNN for depth regression. 
Later, \cite{liu2015deep} combines conditional random field (CRF) with CNN to further improve the quality of estimated depth. 
\cite{fu2018deep} proposed a seminal ordinal regression loss instead of metric $l1/l2$ loss, recasting depth regression as an ordinary regression problem. 
Some works~\cite{qi2018geonet,Yin2019enforcing} introduce extra geometric constraints to improve depth estimation. However, these methods suffer from the scale ambiguity problem due to single-view depth estimation is an ill-posed problem.

\vspace{-0.15in}
\paragraph{Multi-view stereo depth estimation} 
Recently, some learning based methods~\cite{yao2018mvsnet,wang2018mvdepthnet,im2019dpsnet,yao2018mvsnet} based plane-sweeping volume algorithm~\cite{gallup2007real} have achieved promising improvements against their traditional counterparts~\cite{hosni2012fast,hirschmuller2007stereo}.
Some works~\cite{kusupati2020normal,long2020occlusion} introduce surface normal as extra constraint to further improve multi-view depth estimation. 
However, most of the methods are designed for individual images not suitable for videos, thus directly extending these methods from images to video sequence causes flickering artifacts. 
Moreover, these methods mostly adopt fully 3D convolution network for cost regularization, which is memory-consuming, expensive, and slow.
Although one method~\cite{sinha2020depth} tries to avoid 3D convolution layers by replacing plane-sweeping construction by correspondence triangulation, singular value decomposition (SVD) operations used in its triangulation module lead to more time consumption than plane-sweeping methods. 

\vspace{-0.15in}
\paragraph{Temporal coherence in video depth estimation} To utilize temporal coherence to improve depth accuracy, some single image depth estimation methods~\cite{fu2018deep,wang2019recurrent,zhang2019exploiting,patil2020don} exploit recurrent neural units to encode temporal correlation in latent space. 
However, these methods suffer from depth scale ambiguity problem and show poor generalization to frames with unseen camera motions. 
Another single depth estimation method~\cite{luo2020consistent} adopts an online training strategy, performing model re-training on testing videos with a geometric consistency loss function. 
However, the online training scheme costs more than 40 minutes for a video with 244 frames and might suffer from model degradation due to insufficient online training data. 
One multi-view method, neuralrgbd~\cite{liu2019neural} exploits temporal coherence by accumulating depth probability over time. 
However, it can only make use of one preceding estimation to improve the depth estimation of the current frame.
In contrast, our method can take advantage of the temporal coherence of multiple preceding estimations for more accurate and temporally consistent depth maps.

\section{Method}
\label{Method}
We propose an end-to-end pipeline for multi-view depth estimation as shown in Figure~\ref{fig:overview}. 
In the training stage, our model takes a video clip with 5 images $\{I_{t-2},I_{t-1},I_{t},I_{t+1},I_{t+2}\}$ as input, and estimate the depth maps of three target images $\{I_{t-1},I_{t},I_{t+1}\}$ jointly with short-term temporal coherence.
In the inference stage, our model could propagate long-term temporal coherence through the whole video by an Epipolar Spatio-Temporal Memory (ESTM) inference operation.

Briefly, our method has the following four parts, which will be discussed in detail in the subsequent sections:

\renewcommand{\labelenumi}{\roman{enumi}}
\begin{enumerate}
    \item  Hybrid cost volume generation (see Section~\ref{hybrid_vol}). It obtains a regularized matching volume from each target image with its source images using \textit{MatchNet}, and in parallel extracts context feature volume of the target image using \textit{ContextNet}. Finally, it fuses the matching volume and context volume together to a hybrid cost volume.
    \vspace{-0.1in}
    \item Epipolar Spatio-Temporal Transformer (see Section~\ref{transformer}). It is applied on the hybrid cost volumes of all the target images to enforce temporal coherence.
     \vspace{-0.1in}
    \item Depth extraction (see Section~\ref{refinenet}). In this part initial depth maps are first extracted from the transformed cost volumes and further refined by \textit{RefineNet}.
     \vspace{-0.1in}
    \item Epipolar Spatio-Temporal Memory (ESTM) inference operation (see Section~\ref{ESTM}). In the inference stage, this operation propagates long-term temporal coherence through the whole video.
\end{enumerate}

\subsection{Hybrid cost volume generation}
\label{hybrid_vol}
For computational efficiency, we utilize two expert sub-networks, \textit{MatchNet} and \textit{ContextNet}, to learn two types of cost volumes, 3D matching volume and 2D context volume, respectively.

\subsubsection{Matching volume generation}
\label{matchnet}
To simplify the exposition, we take frame $I_{t}$ as an example to construct its matching volume. 
Frame $I_{t}$ and its neighboring frames $I_{t-1}, I_{t+1}$ are taken as reference image and source images respectively, since they share the largest overlapping regions. 
Without loss of generality, here we only consider the reference image and one of its source images, denoted as $I_{r}$ and $I_{s}$ respectively. 

\textbf{Feature extraction} We first pass $I_{r}$ and $I_{s}$ through a spatial pyramid pooling (SPP) \textit{feature extractor}~\cite{chang2018pyramid} to extract corresponding hierarchical feature maps $F_{r}$ and  $F_{s}$. 
For computational efficiency, the feature maps are down-sampled by four to the original image size with 32 channels.

\textbf{Raw matching volume} A raw matching volume for reference image $I_{r}$ is constructed by backprojecting source feature map $F_{s}$ into the coordinate system defined by $I_{r}$ at a stack of fronto-parallel virtual planes.
The virtual planes are uniformly sampled over a range of depths $z_1, z_2, \cdots ,z_D$, which are chosen to span the ranges observed in the dataset. Following prior works~\cite{im2019dpsnet,long2020occlusion,liu2019neural,kusupati2020normal,wang2018mvdepthnet}, we set $D=64$ in all experiments.
The coordinate mapping from the source feature map $F_{s}$ to the reference feature map $F_{r}$, at each depth $z_{m}$, is determined by a planar homography transformation:
\begin{equation}
u'_{m}\sim H_{m}u,u'_{m}\sim \mathbf{K}[{\mathcal{R}}_{s}|{\mathbf{t}}_{s}]\begin{bmatrix}
({\mathbf{K}}^{-1}u)z_{m} \\
1 
\end{bmatrix}
\label{mapping}
\end{equation}
where the '$\sim$' denotes the projective equality, $u$ is homogeneous coordinate of a pixel in the reference image, $u'_{m}$ is the projected homogeneous coordinate of $u$ on the paired source feature map. $K$ denotes the intrinsic parameters of the camera, $\{ \mathcal{R}_{s}, \mathbf{t}_{s} \}$ are the rotation and the translation of the source image ${I}_{s}$ relative to the reference image $I_{r}$.

Based on the above mapping, we warp the source feature map into all the virtual planes to construct a feature volume with dimension $C \times D \times H \times W$. 
Finally, we concatenate the feature volume with $F_r$ at each virtual plane to increase the dimension to $2C \times D \times H \times W$. 
By concatenation operation, the network could receive necessary information to perform feature matching between $F_r$ and $F_s$ without decimating the feature dimension~\cite{wang2018mvdepthnet,long2020occlusion}.
With $N$ source images, we will obtain $N$ raw matching volumes.

\textbf{MatchNet} 
The $N$ raw matching volumes are first processed by three 3D convolution layers to decrease their dimension to $C \times D \times H \times W$. Then a view average pooling operation is performed on the volumes to aggregate information across different source images, yielding a single aggregated volume. Finally, the aggregated volume is further regularized by a series of 3D convolution layers.
Note that our \textit{MatchNet} is only responsible for learning local features for matching. We will use another network for learning global context information to complement \textit{MatchNet} for more efficient depth estimation, as will be discussed in the next section.

\subsubsection{Context volume generation}
\label{contextnet}
Prior methods~\cite{liu2019neural,im2019dpsnet,kusupati2020normal} adopt a heavy 3D regularization networks to learn 2D context information together with 3D local matching clues in a mixed manner, and the network is made very deep for increased ability of learning the two types of information.
We observe that the global context information is essentially 2D information, so it is unnecessary to use a 3D network to learn it. 
Hence, we decouple the global context information from the local matching information and use a 2D network, called {\em \textit{ContextNet}}, to learn the former. This makes the network simpler and, consequently, more efficient to train and run.


Specifically, we use Resnet-50 as the \textit{ContextNet}. The output of \textit{ContextNet} is a learned feature volume with size $C' \times H \times W$, where $C'$ is the number of feature channels and the same as the number of virtual planes $D$. To fuse 3D matching volume and the 2D context volume, we expand the dimensions of context volume to $1 \times D \times H \times W$. Finally we concatenate the regularized matching volume and the expanded context volume together to get a hybrid cost volume, with size $(C+1) \times D \times H \times W$. 
As shown in Figure~\ref{fig:overview}, we repeatedly apply the hybrid cost volume generation operation for images $I_{t-1}, I_{t}, I_{t+1}$ and obtain three corresponding hybrid cost volumes $\mathbf{C}_{t-1},\mathbf{C}_{t},\mathbf{C}_{t+1}$.

\subsection{Epipolar Spatio-Temporal transformer}
\label{transformer}
To associate temporal coherence with the three hybrid cost volumes $\mathbf{C}_{t-1},\mathbf{C}_{t},\mathbf{C}_{t+1}$, we propose a novel Epipolar Spatio-Temporal (EST) transformer.

\textbf{Consistency constraint} Our EST transformer is inspired by the photometric consistency assumption: a 3D point in world space will be projected into visible images $I_{t-1},I_{t},I_{t+1}$, and the image textures near their projections should bear high similarity. 
We formulate depth estimation as an occupancy estimation problem: if a pixel $(u,v)$ of image $I_{t}$ has depth value $d$ then the voxel $(u,v,d)$ in $\mathbf{C}_{t}$ is occupied, that is, the learned features of $\mathbf{C}_{t}$ encode the probability of occupancy for each voxel. The hybrid cost volumes are treated as multiple occupancy measurements for the same 3D world space in different viewpoints, namely, for a 3D point in world space, its corresponding voxels of the volumes $\mathbf{C}_{t-1},\mathbf{C}_{t},\mathbf{C}_{t+1}$ should keep similar embedding vectors.

\begin{figure}[t]
\setlength{\abovecaptionskip}{0pt}
\setlength{\belowcaptionskip}{0pt}
    \centering
    \adjincludegraphics[width=0.8\linewidth]{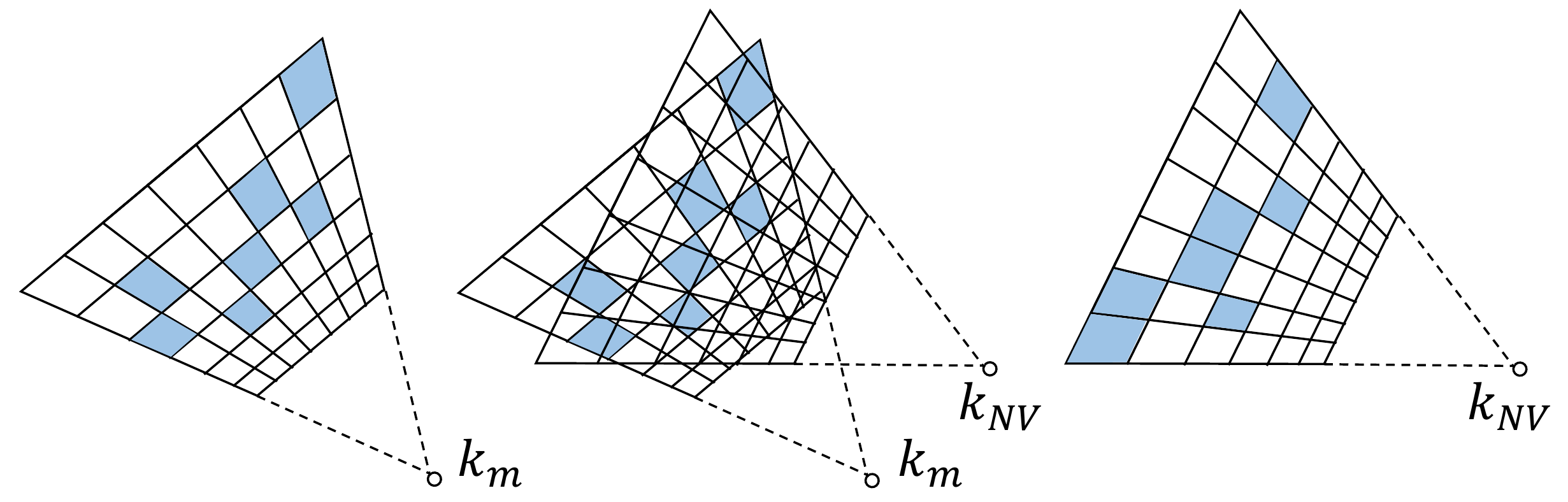}
    \caption{Epipolar warping. To apply EST transformer on the query volume $\mathbf{C}_{q}$, we need to resample the memory keys and memory values with respect to a novel view, the camera view of query volume. For better visualization, we only look at a planar slice of the key volumes. Each voxel of the volumes is encoded by a one-dimension feature, and here we only colorize some voxels for simplification. }
    \label{fig:volume_resample}
\end{figure}

\textbf{Epipolar warping} To associate temporal coherence with the hybrid volumes, we should first perform epipolar warping (see Figure~\ref{fig:volume_resample}) to convert the hybrid volumes into the same camera coordinate space. Assuming a 3D point $(X_{w}, Y_{w}, Z_{w})$ in the world coordinate is observed by target images $I_{t}$ and $I_{t+1}$, that is, the voxel $(u,v,d)$ of $\mathbf{C}_{t}$ and $(u',v',d')$ of $\mathbf{C}_{t+1}$ are occupied. The coordinate mapping from $(u,v,d)$ to $(u',v',d')$ can be easily derived from Equation~\ref{mapping}. Using this mapping, we  warp $\mathbf{C}_{t-1}$ and $\mathbf{C}_{t+1}$ into the camera coordinate space of $\mathbf{C_{t}}$ and obtain two warped hybrid volumes $\mathbf{C}_{t-1}^{warp}, \mathbf{C}_{t+1}^{warp}$. After converted into the same camera coordinate, the two warped volumes and $\mathbf{C}_{t}$ should contain similar features in voxels for overlapped regions.

\begin{figure}[t]
\setlength{\abovecaptionskip}{0pt}
\setlength{\belowcaptionskip}{0pt}
    \centering
    \includegraphics[width=\columnwidth]{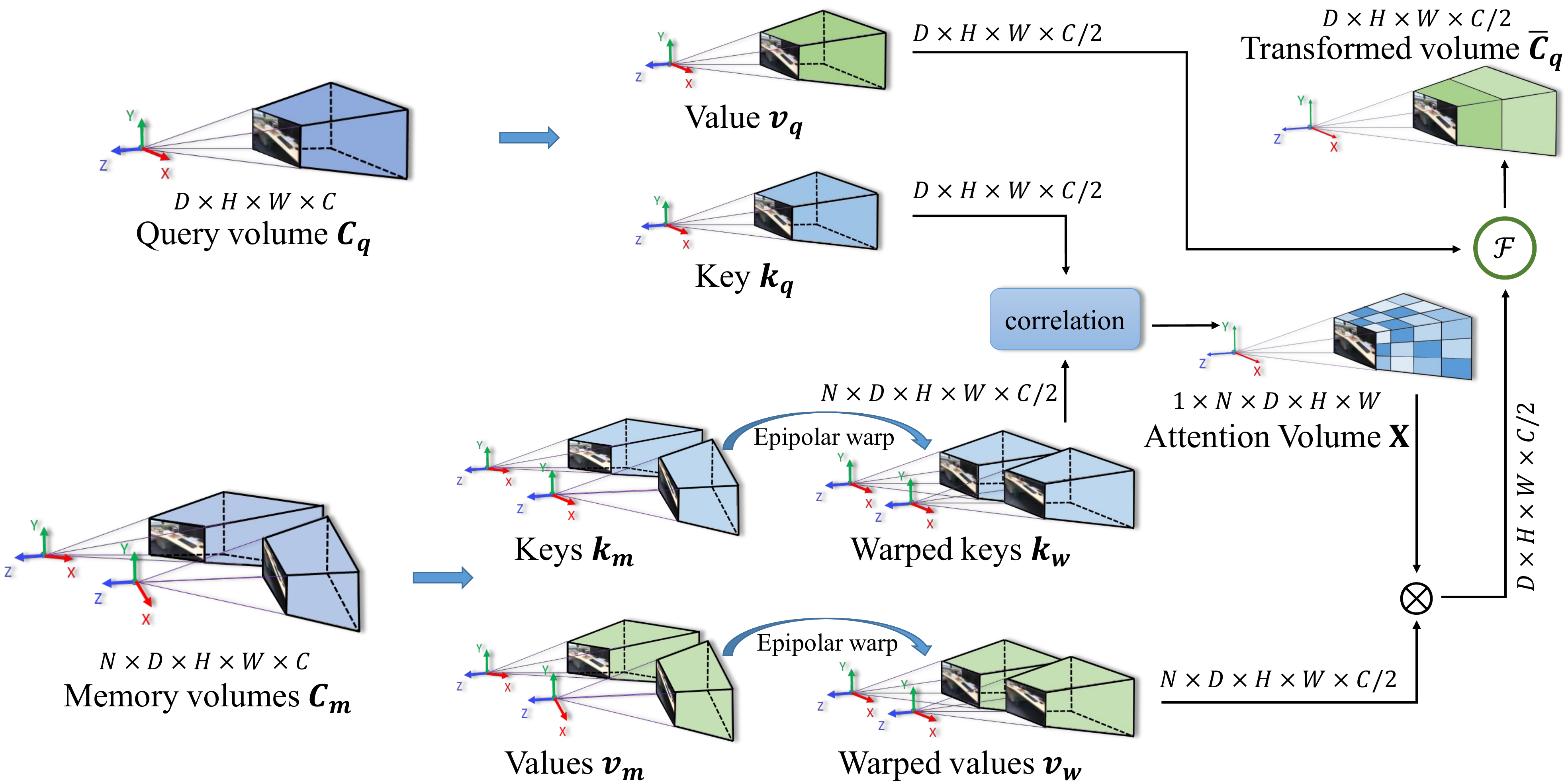}
    \caption{The structure of EST Transformer. Both query volume and memory volumes are encoded into pairs of \textbf{key} and \textbf{value} maps through shared convolution layers. The memory keys and memory values are epipolar warped into the camera space of the query volume. Similarities between query key and warped memory keys are computed to determine when-and-where to retrieve relevant memory values from. 
    Finally we adaptively fuse the retrieved values with query value together for final output. Here $\otimes$ means matrix inner product, and $\mathcal{F}$ denotes fusion function.}
    \label{fig:transformer}
\end{figure}

\textbf{EST transformer} As depicted in Figure~\ref{fig:transformer}, we denote the hybrid volume to be transformed $\mathbf{C}_{t}$ as query volume and others ($\mathbf{C}_{t-1}$ and $\mathbf{C}_{t+1}$) as memory volumes. 
For computational efficiency, instead of applying EST transformer on the hybrid volumes, we first feed the query volume and memory volumes into two parallel and identical convolution layers to generate two new squeezed feature maps \textbf{key}s $k \in \mathbb{R}^{C/2 \times D \times H \times W}$ and \textbf{value}s $v \in \mathbb{R}^{C/2 \times D \times H \times W}$.
The memory keys and memory values denoted as $k_{m}$ and $v_{m}$, are first epipolar warped into the camera space of $\mathbf{C_{t}}$, obtaining warped keys and values denoted as $k_{w}$ and $v_{w}$. Then we calculate the correlation between the query key $k_{q}$ and the warped keys $k_{w}$, yielding a correlation volume, which measures the similarity of $k_{q}$ and $k_{w}$. 
Finally we apply a softmax layer to get an attention volume $\mathbf{X} \in \mathbb{R}^{1 \times N \times D \times H \times W}$:
\begin{equation}
    x_{i} = \frac{exp(k_{q}\cdot k_{w}^{i})}{\sum_{i=1}^{N}exp(k_{q}\cdot k_{w}^{i})}
\end{equation}
where $x_i \in \mathbb{R}^{1 \times 1 \times D \times H \times W}$ measures the similarity of query to the warped key of $i^{th}$ memory volume, $N$ is the number of memory volumes, and $\cdot$ means dot product. 

The attention map $\mathbf{X}$ is used to retrieve relevant values from all warped memory values $v_{w}^{i},i=1,\cdots,N$. Finally we fuse the query value and retrieved values together to obtain the final output $\mathbf{\bar{C}}_{q}$:
\begin{equation}
    \mathbf{\bar{C}}_{q}=f(v_{q}, \sum_{i=1}^{N}x_{i} v_{w}^{i} )
\end{equation}
where $v_{w}^{i}$ is the $i^{th}$ warped memory value, and $f(\cdot,\cdot)$ denotes a fusion function.

\begin{figure*}[t]
\setlength{\abovecaptionskip}{0pt}
\setlength{\belowcaptionskip}{0pt}
    \centering
    \includegraphics[width=0.9\textwidth]{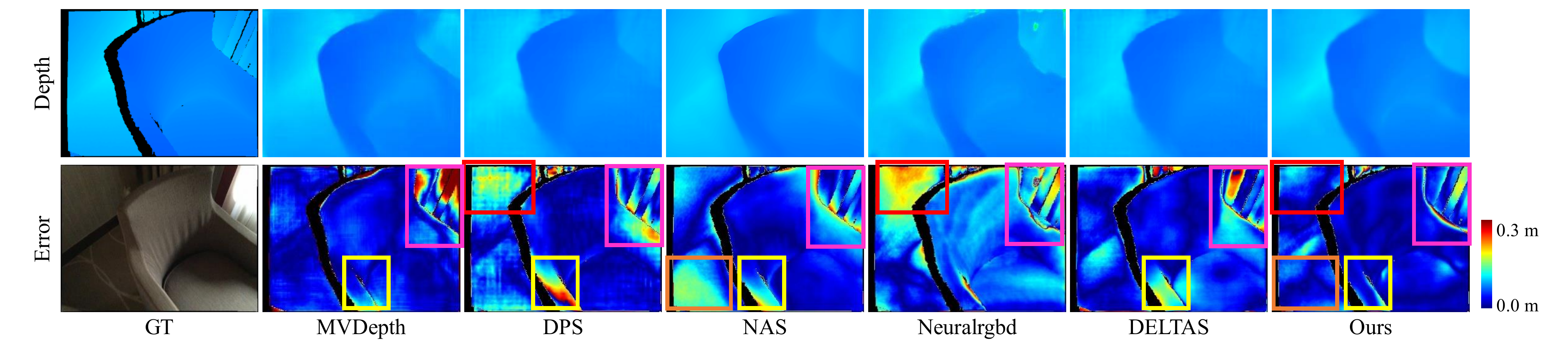}
    \caption{Depth comparison with the SOTAs. Our estimated depth map is more accurate and has less noises even in texture-less regions (the marked boxes). Black regions are invalid regions where gt depth values are missed. More comparisons are in the supplementary materials.}
    \label{fig:depth_compare}
\end{figure*}

We explore two types of fusion function: concatenation fusion and adaptive fusion. For concatenation fusion, we just simply concatenate  query value and the retrieved values together, which is straightforward and introduces no trainable parameters. However, due to occlusion or surface reflection, there might exist incorrect information in retrieved values but the concatenation operation integrates all information equally. To avoid propagating wrong information to current estimation, we propose an adaptive fusion operation to fuse the query value and the retrieved values:
\begin{equation}
    f(v_{q},y)=w\odot y+(1-w)\odot g(v_{q}, r\odot y), y = \sum_{i=1}^{N}x_{i} v_{w}^{i}
\end{equation}
\LXX{where $\odot$ means Hadamard product, $w,r \in \mathbb{R}^{D \times H \times W}$ are two learned weight volumes which measure the reliability of the retrieved values, $g(\cdot, \cdot)$ is a convolution layer.
Unless otherwise specified, EST transformer refers to the adaptive transformer. As shown in Table~\ref{tab:scannet_depth_eval}, the adaptive fusion outperforms the simple concatenation fusion. }

The EST transformer is applied to the other two hybrid cost volumes $\mathbf{C}_{t-1}$ and $\mathbf{C}_{t+1}$, yielding three transformed cost volumes $\mathbf{\bar{C}}_{t-1}, \mathbf{\bar{C}}_{t}$ and $\mathbf{\bar{C}}_{t+1}$,
We first apply two convolution layers on the transformed volumes to produce intermediate volumes with size $D \times H \times W$, and then utilize the softmax operator over the depth dimension, so that we obtain probability volumes $\mathbf{P}$.

\subsection{RefineNet and depth regression}
\label{refinenet}
We extract a depth map from probability volume $\mathbf{P}$ by \textit{soft argmax} operation~\cite{kendall2017end}, which calculates the expected depth. 
We denote the depth map regressed from probability volume as initial depth map, since its size is downsized by four compared with original image size, whose fine grained features are lost and boundary edges are jagged. 
So we use a two-stage \textit{RefineNet} to gradually upsample the initial depth maps and enhance fine-grained features, yielding a $1/2$ resolution depth map and a full-resolution depth map. Besides, we also extract a depth map from hybrid cost volumes before being applied EST transformer. 
Consequently, we obtain estimated depth maps of four stages, and we denote the four types of depth map from hybrid cost volumes, transformed cost volumes, and two-stage \textit{RefineNet} as $\mathbf{D}_{s}, s=0,1,2,3$.

Unlike single-view depth loss used in prior works, we utilize multi-view depth loss to provide multiple supervision signals in different viewpoints:
\begin{equation}
    loss=\frac{1}{N} \sum_{s=0}^{3} \sum_{i=1}^{N} \lambda^{s-3} \left\| \mathbf{D}_{s}^{i}-\hat{\mathbf{D}}_{s}^{i}\right\|_{1}
\end{equation}
where $i$ is the the index of target image, $N$ is the number of target images, $\hat{D}$ means groud truth depth map, and $\left\| \cdot \right\|_{1}$ means L1 norm. The weight $\lambda$ is set to 0.8 in all experiments. 

\begin{figure}[t]
\setlength{\abovecaptionskip}{0pt}
\setlength{\belowcaptionskip}{0pt}
    \centering
    \includegraphics[width=\linewidth]{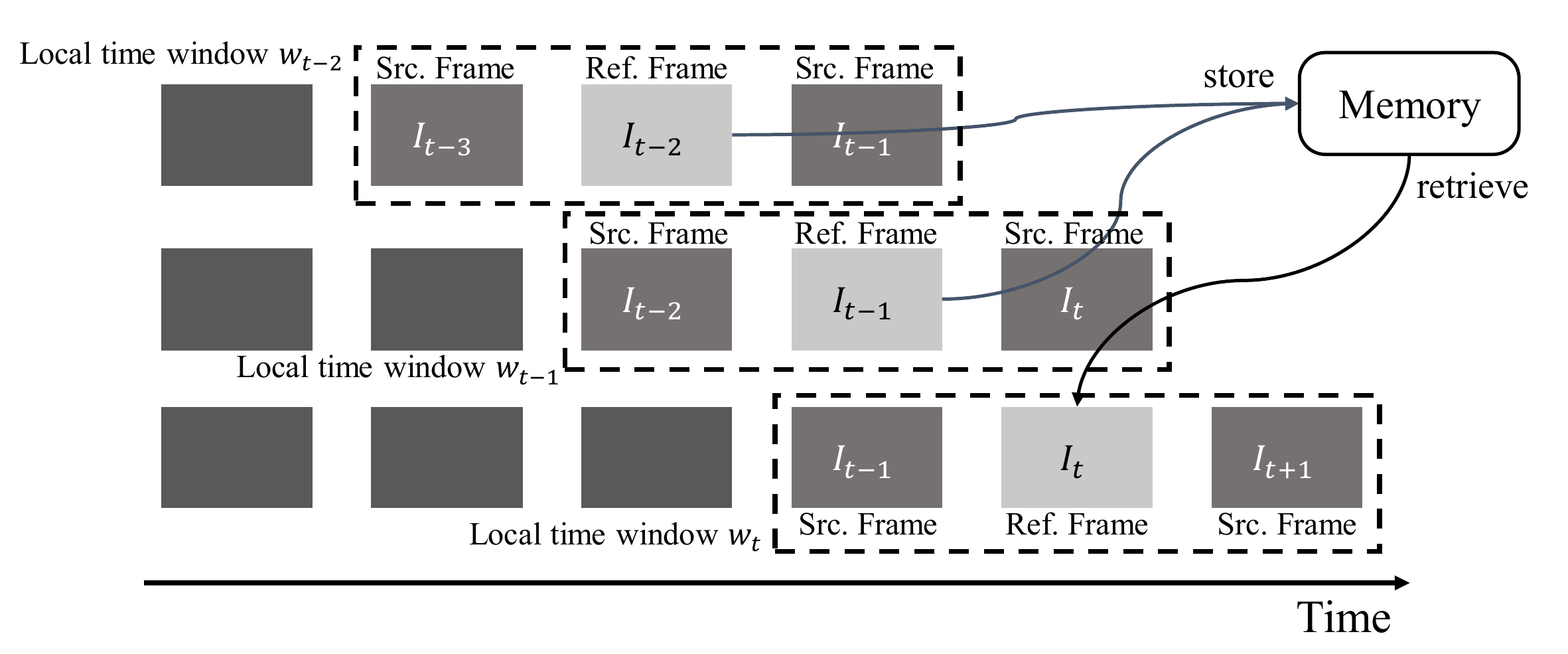}
    \caption{Epipolar Spatio-Temporal Memory inference operation.}
    \label{fig:ESTM}
\end{figure}

\subsection{Epipolar Spatio-Temporal Memory Inference}
\label{ESTM}
In the training stage, our model takes a short video sequence with 5 frames as input and jointly estimate the depth maps of three target images with short-term temporal coherence.
To propagate long-term temporal coherence through the whole video, we propose an Epipolar Spatio-Temporal Memory (ESTM) inference operation. 
As depicted in Figure~\ref{fig:ESTM}, we hold a sliding window containing one reference image and two source images to estimate the depth map of the current frame $I_t$. 
Using the EST transformer, we retrieve relevant values from a memory space storing the pairs of keys and values of $N$ past frames, thus useful information at different space-time locations can be utilized for estimating the depth map of the current frame.
When the sliding window moves on, the memory space will be also updated accordingly, by which operation the long-term temporal coherence is propagated through the whole video.

\section{Datasets and Implementation details}
\noindent
\textbf{Datasets} We use ScanNet dataset~\cite{dai2017scannet} for training our end-to-end pipeline. The whole dataset consists of more than 1600 indoor scenes, which provides color images, ground truth depth maps and camera poses. 
Some MVS methods, DPS~\cite{im2019dpsnet} and MVDepth~\cite{wang2018mvdepthnet} are trained on DeMoN~\cite{ummenhofer2017demon} dataset, but DeMoN dataset mainly consists of two-view image pairs, which is not appropriate for our setting. 
For a fair comparison, we finetune DPS and MVDepth on ScanNet, and evaluate all methods on the official test split of ScanNet.

Furthermore, we test our method on  7scenes~\cite{shotton2013scene} and SUN3D~\cite{xiao2013sun3d} datasets for cross-dataset evaluation. 
7scenes and SUN3D datasets also provide color images, gt depth maps and camera poses. Unlike prior works~\cite{ummenhofer2017demon,im2019dpsnet,kusupati2020normal} sample two-view image pairs from videos, we directly adopt whole videos for evaluation.

\noindent
\textbf{Implementation} In the training stage, we use a video clip with 5 frames as input. The frames are sampled from 30fps video with an interval of 10. Our model is implemented by Pytorch with Adam optimizer ($lr= 0.00004$, $\beta_{1}=0.9$, $\beta_{2}=0.999$, $weight\_decay = 0.00001$). We train our model for 7 epochs (~115k iterations) with batch size 4 on four GeForce RTX 2080 Ti GPUs, and the learning rate is downscaled by a factor of 2 every two epochs.

\section{Experiments}
We compare our method with SOTA models in two aspects: depth accuracy and computational efficiency.
\subsection{Evaluation metrics}
To quantitatively evaluate the estimated depth, we use the standard metrics defined in \cite{eigen2014depth}: i) inlier ratios ($\sigma < {1.25}^{i} \text{ where } i\in\{1,2,3\}$);
ii) Absolute error (Abs); iii) Absolute relative error (Abs Rel); iv) Square Relative error (Sq Rel); iv)  Root Mean Square Error (RMSE); v) RMSE in log space (RMSE log). 

To measure computational complexity of models, we adopt the following metrics: i) the number of trainable parameters; ii) the number of Multiply–Accumulate Operations (MACs); iii) inference memory usage; iv) inference time cost. 

\begin{table*}[t]
\begin{center}
\caption{Depth comparison over ScanNet~\cite{dai2017scannet} and 7scenes~\cite{shotton2013scene} datasets. Our method outperforms the other methods by a large margin. 
We report results in two depth ranges, since Neuralrgbd is trained in range $0 \sim 5m$. 
Complete tables are in the supplementary materials.
}
\label{tab:scannet_depth_eval}
\resizebox{0.9\linewidth}{!}{%
\begin{tabular}{c l | c c c c c | c c c c c}
\hline
\multirow{2}{*}{Range}& \multirow{2}{*}{Method} & \multicolumn{5}{c|}{ScanNet} & \multicolumn{5}{c}{7scenes} \\
&  & Abs Rel & Abs & Sq Rel & RMSE  & $\sigma < 1.25$ & Abs Rel & Abs & Sq Rel & RMSE  & $\sigma < 1.25$ \\
\hline
\multirow{9}{*}{10m} & MVDepth~\cite{wang2018mvdepthnet} & 0.1167 & 0.2301 & 0.0596 & 0.3236 & 84.53 &  0.2213 & 0.4055 & 0.2401 & 0.5154  & 67.33 \\
& MVDepth-FT & 0.1116 & 0.2087 & 0.0763 & 0.3143 & 88.04 & 0.1905 & 0.3304 & 0.1319 & 0.4260  & 71.93 \\
&DPS~\cite{im2019dpsnet} & 0.1200 & 0.2104 & 0.0688 & 0.3139  & 86.40  & 0.1963 & 0.3471 & 0.1970 & 0.4625 & 72.51 \\
&DPS-FT & 0.0986 & 0.1998 & 0.0459 & 0.2840 & 88.80 &  0.1675 & 0.2970 & 0.1071 & 0.3905 & 76.03 \\
&NAS~\cite{kusupati2020normal} & 0.0941 & 0.1928 & 0.0417 & 0.2703 & 90.09 & 0.1631 & 0.2885 & 0.1023 & 0.3791  & 77.12 \\
&CNM~\cite{long2020occlusion} & 0.1102 & 0.2129 & 0.0513 & 0.3032  & 86.88 & 0.1602 & 0.2751 & 0.0819 & 0.3602 & 76.81\\
&DELTAS~\cite{sinha2020depth} & 0.0915 & 0.1710 & 0.0327 & 0.2390 & 91.47  & 0.1548 & 0.2671 & 0.0889 & 0.3541 & 79.66\\
& Ours-EST(concat) & 0.0818 & 0.1536 & 0.0301 & 0.2234 & 92.99 &  \textbf{0.1458} & 0.2554 & 0.0745 & 0.3436  & 79.82 \\
&Ours-EST(adaptive) & \textbf{0.0812} & \textbf{0.1505} & \textbf{0.0298} & \textbf{0.2199} & \textbf{93.13} & \underline{0.1465} & \textbf{0.2528} & \textbf{0.0729} & \textbf{0.3382}  & \textbf{80.36}\\
\hline
\multirow{3}{*}{5m}&Neuralrgbd~\cite{liu2019neural} & 0.1013 & 0.1657 & 0.0502 & 0.2500 & 91.60 & 0.2334 & 0.4060 & 0.2163 & 0.5358  & 68.03\\
&Ours-EST(concat) & 0.0811 & 0.1469 & 0.0279 & 0.2066 & 93.19 & \textbf{0.1458} & 0.2554 & 0.0745 & 0.3435 & 79.82\\
&Ours-EST(adaptive) & \textbf{0.0805} & \textbf{0.1438} & \textbf{0.0275} & \textbf{0.2029} & \textbf{93.33} & \underline{0.1465} & \textbf{0.2528} & \textbf{0.0729} & \textbf{0.3382}  & \textbf{80.36} \\
\hline
\end{tabular}%
}
\end{center}
\vspace{-4mm}
\end{table*}

\subsection{Depth evaluation}
\begin{figure}[t]
\setlength{\abovecaptionskip}{0pt}
\setlength{\belowcaptionskip}{0pt}
    \centering
    \includegraphics[width=0.9\columnwidth]{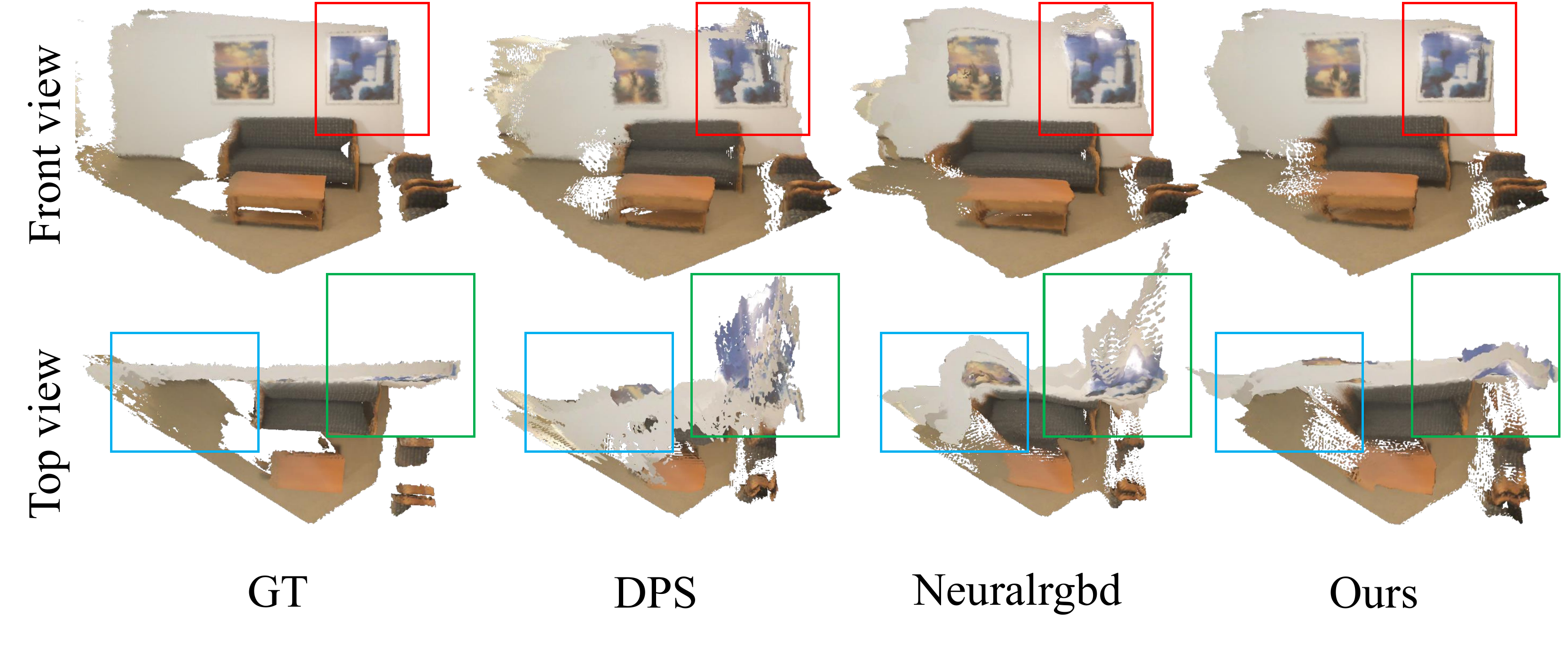}
    \caption{3D reconstruction results by TSDF fuison method~\cite{zeng20163dmatch}. To evaluate temporal coherence, we fuse the estimated depth maps of 10 successive sampled video frames with the interval of 10. Compared to DPS~\cite{im2019dpsnet} and Neuralrgbd~\cite{liu2019neural}, our method generates more faithful and temporally consistent depth maps.}
    \label{fig:reconstruction}
\end{figure}

\paragraph{Depth accuracy} Since MVDepth~\cite{wang2018mvdepthnet} and DPS~\cite{im2019dpsnet} are trained on DeMon dataset~\cite{ummenhofer2017demon} not on ScanNet~\cite{dai2017scannet} dataset.  
For fair comparisons, we fine-tune MVDepthNet and DPSNet on ScanNet. 
As shown in Table~\ref{tab:scannet_depth_eval}, our model significantly outperforms all other methods over both ScanNet and 7scenes datasets. Compared with Neuralrgbd~\cite{liu2019neural}, our model not only achieves better results in ScanNet dataset but also shows superior generalization ability on unseen 7scenes dataset.
In Figure~\ref{fig:depth_compare}, compared with DPS~\cite{im2019dpsnet}, NAS~\cite{kusupati2020normal}, Neuralrgbd~\cite{liu2019neural} and DELTAS~\cite{sinha2020depth}, our estimated depth maps are more accurate and contain fewer noises. 
Our depth maps have smaller errors in texture-less regions, such as floor, wall, and sofa. 
Furthermore,  we feed the estimated depth maps into TSDF fusion system for 3D reconstruction, as shown in Figure~\ref{fig:reconstruction}, our reconstructed model is more accurate and has fewer outliers.


\vspace{-4mm}
\paragraph{Temporal coherence} To measure the temporal consistency of the estimated depth maps, we adopt standard deviation of the mean absolute error of the estimated depth maps for evaluation. As shown in Table~\ref{tab:temporal_compare}, the estimated depth maps of our methods are more accurate and temporally consistent than those of other methods (See supplementary materials for more details). 

\begin{table}[!htp]
\begin{center}
\caption{Comparison of temporal coherence over ScanNet dataset with depth evaluation range $0 \sim 5m$. 
}
\label{tab:temporal_compare}
\resizebox{\linewidth}{!}{%
\begin{tabular}{l | c c  c c c }
\hline
Metric & DPS\cite{im2019dpsnet} & NAS~\cite{kusupati2019normal} & Neuralrgbd~\cite{liu2019neural} & DETALS~\cite{sinha2020depth} & Ours \\
\hline
Abs & 0.1887 & 0.1823 & 0.1642 & 0.1650 & \textbf{0.1432}\\
Std & 0.2243 & 0.2177 & 0.1848 & 0.1886 & \textbf{0.1673}\\

\hline
\end{tabular}
}
\end{center}
\vspace{-4mm}
\end{table}

\subsection{Analysis of computation complexity }
To evaluate the computational efficiency of our model, we compare it with three plane-sweeping based methods, namely DPS, NAS, and neuralrgbd, plus the correspondences triangulation based method DELTAS.
We run the models on one RTX 2080Ti GPU with the same setting: one reference image with two source images with a size of $320 \times 256$. We run our model using ESTM inference operation with 2 memory volumes.

Table~\ref{tab: memory_analysis} shows that DELTAS has the most trainable parameters but the lowest MACs, owing to the low computational cost of a fully 2D convolution network it uses. 
However, DELTAS consumes much more time than all the other plane-sweeping stereo methods because its correspondence triangulation module needs to perform time-consuming Singular Value Decomposition (SVD). 
Our method achieves significantly faster speed than the other methods. It takes about 40 milliseconds for our model to perform the adaptive ESTM operation, and 31 milliseconds to forward convolution layers, thus 71 milliseconds in total. 

\begin{table}[tp]
\setlength{\abovecaptionskip}{0pt}
\setlength{\belowcaptionskip}{0pt}
\begin{centering}
\caption{Memory and computation complexity analysis.}
\resizebox{\columnwidth}{!}{%
\begin{tabular}{l c c c c}
\hline
Model & Params & MACs & Memory & Time  \\
\hline
DPS~\cite{im2019dpsnet}  & \textbf{4.2M} & 442.7G & \textbf{1595M} & 337ms \\
NAS~\cite{kusupati2020normal} & 18.0M & 527.7M & 1689G & 212ms \\
Neuralrgbd~\cite{liu2019neural} & \underline{5.3M} & 616.6G & 2027M & 195ms \\
DELTAS~\cite{sinha2020depth}  & 124.6M & \textbf{98.6G} & 2395M & 495ms \\
Ours-ESTM & 36.2M & 176.9G & 1799M & \textbf{71ms} \\
\hline
\end{tabular}
}

\label{tab: memory_analysis}
\end{centering}
\vspace{-4mm}
\end{table}

\begin{figure*}[htp]
\setlength{\abovecaptionskip}{0pt}
\setlength{\belowcaptionskip}{0pt}
    \centering
    \includegraphics[width=0.9\textwidth]{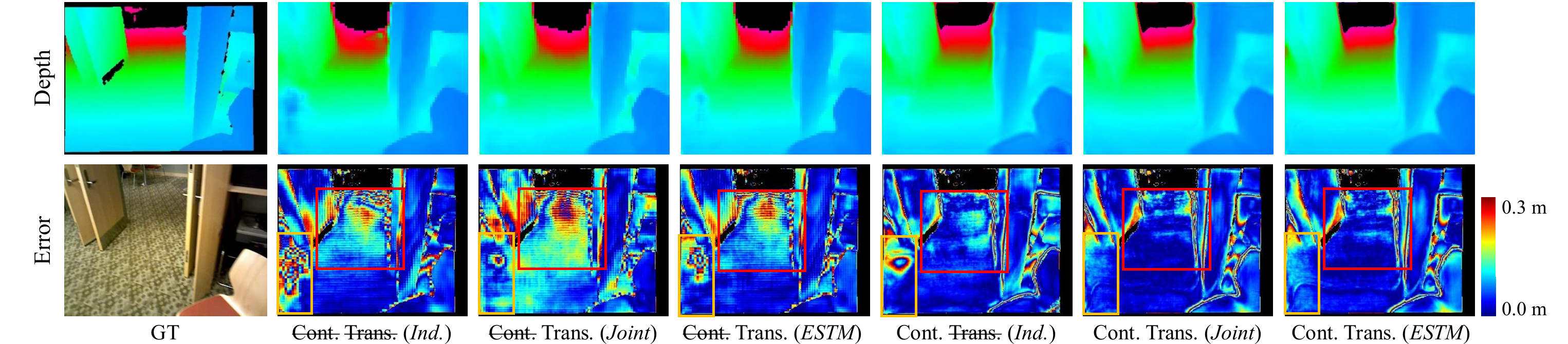}
    \caption{Effects of EST transformer and \textit{ContextNet}. Here \textit{\st{Cont.}} denotes models without \textit{ContextNet}, \textit{\st{Trans.}} denotes models without EST transformer, \textit{Ind.} means that independently estimate depth maps by the model without EST transformer, \textit{Joint} means that jointly estimate multiple depth maps at a time, and \textit{ESTM} means that estimate depth maps sequentially by our ESTM inference operation.}
    \label{fig:ablation}
    \vspace{-2mm}
\end{figure*}






\subsection{Ablation studies}
In this section we evaluate the efficacy of our EST transformer and hybrid cost regularization network.

\textbf{Epipolar Spatio-Temporal transformer} 
We consider several variants of our method for ablation studies.
We denote the depth estimated by the model without EST transformer as \textit{independent depth}, the depth jointly estimated by the model with EST transformer as \textit{joint depth}, and the depth sequentially estimated by the model using ESTM operation as \textit{ESTM depth}.
As shown in Table~\ref{tab: ablation}, when using the hybrid cost regularization network, both \textit{joint depth} and \textit{ESTM depth} are better than \textit{independent depth}.
However, when adopting the pure 3D regularization network (without \textit{ContextNet}), \textit{joint depth} is worse than \textit{independent depth}, and the improvement of \textit{ESTM depth} is trivial.

That is because if the estimated depth is not accurate enough, the joint estimation will propagate wrong information across the multiple depth maps. But ESTM suffers from it less because the errors can be alleviated gradually as more frames are processed sequentially. Overall, the combination of the hybrid regularization network and EST transformer can boost the best performance. Detailed discussions are included in the supplementary materials.

Furthermore, we compare our EST transformer with the recurrent neural network. We replace EST transformer by Gated Recurrent Units (GRU) in our model for comparison. As shown in Table~\ref{tab: trans_vs_rnn}, models with EST transformer outperforms the model with RNN.

\textbf{Efficacy of ContextNet} As shown in Table~\ref{tab: ablation}, our models with \textit{ContextNet} significantly outperform those without \textit{ContextNet} in all inference types: independently, jointly, and by ESTM operation. In Figure~\ref{fig:ablation}, the depth maps estimated by the models with \textit{ContextNet} have smaller errors in pattern-rich regions (e.g. floor) and texture-less regions (e.g. pillar) than those without \textit{ContextNet}. 
We show that our \textit{ContextNet} is indispensable and provides complementary and important information for cost regularization.

\begin{table}[tp]
\setlength{\abovecaptionskip}{0pt}
\setlength{\belowcaptionskip}{0pt}
\begin{centering}
\caption{The usefulness of \textit{ContextNet} and epipolar transformer. We test models with various settings on SUN3D~\cite{xiao2013sun3d}.}
\resizebox{\linewidth}{!}{%
\begin{tabular}{c c c c c c c}
\hline
Cont.  & Trans. & Inference type& Abs & Sq Rel & RMSE   & $\sigma < 1.25$   \\
\hline
\textcolor{red}{\xmark} & \textcolor{red}{\xmark} & Independent & 0.3333 & 0.0994 & 0.4897 & 80.89\\
\textcolor{red}{\xmark} & \textcolor{green}{\cmark} & Joint & 0.3429 & 0.1291 & 0.4927 & 81.36 \\
\textcolor{red}{\xmark} & \textcolor{green}{\cmark} & ESTM & 0.3319 & 0.1073 & 0.4822 & 81.43 \\
\textcolor{green}{\cmark} & \textcolor{red}{\xmark} & Independent &  0.3220 & 0.0897 & 0.4657 & 82.82\\
\textcolor{green}{\cmark} & \textcolor{green}{\cmark} & Joint & \textbf{0.3133} & \textbf{0.0883} & \textbf{0.4556}  & \textbf{83.52} \\
\textcolor{green}{\cmark} & \textcolor{green}{\cmark} & ESTM & 0.3137 & 0.0884 & 0.4554 & 83.43 \\ 
\hline
\end{tabular}
}
\label{tab: ablation}
\end{centering}
\vspace{-4mm}
\end{table}

\textbf{Memory size of ESTM} 
Using the ESTM inference operation, we can retrieve useful information from an arbitrary number of preceding estimations stored in \textit{memory} space to enhance the estimation of the current frame. 
As shown in Table~\ref{tab: memory_size}, the quality of estimated depth will be improved gradually with the increase of the number of stored preceding estimations, but the improvement becomes less significant with more than 2 stored estimations.
Hence we set the optimal memory size to be 2, as the best trade-off between performance and computational efficiency.

\begin{table}[tp]
\setlength{\abovecaptionskip}{0pt}
\setlength{\belowcaptionskip}{0pt}
\begin{centering}
\caption{The effect of different size of ESTM memory space. Experiments are done over 7scenes~\cite{shotton2013scene}.}
\resizebox{0.95\linewidth}{!}{%
\begin{tabular}{c c c c c c}
\hline
Memory size  & Abs Rel & Abs & Sq Rel & RMSE   & $\sigma < 1.25$   \\
\hline
1 & 0.1530 & 0.2632 & 0.783 & 0.3494 & 79.07 \\
2 & 0.1465 & 0.2528 & 0.0729 & 0.3382 & 80.36\\
3 & \textbf{0.1460} & \textbf{0.2520} & \textbf{0.0727} & \textbf{0.3376} & \textbf{80.44}\\
4 & 0.1461 & 0.2521 & 0.0728 & 0.3377 & 80.44\\
\hline
\end{tabular}

}

\label{tab: memory_size}
\end{centering}
\end{table}

\begin{table}[tp]
\setlength{\abovecaptionskip}{0pt}
\setlength{\belowcaptionskip}{1pt}
\begin{centering}
\caption{EST transformer vs RNN. We replace the EST transformer with Gated Recurrent Units in our model.}
\resizebox{0.8\linewidth}{!}{%
\begin{tabular}{l | c c | c c}
\hline
\multirow{2}{*}{Model} & \multicolumn{2}{c}{ScanNet} & \multicolumn{2}{|c}{SUN3D} \\
  & Abs & $\sigma < 1.25$  & Abs & $\sigma < 1.25$ \\
\hline
Ours-RNN & 0.1680 & 92.67 & 0.3401 & 82.33 \\
Ours-ESTM & \textbf{0.1505} & \textbf{93.13} & \textbf{0.3137} & \textbf{83.52}\\
\hline
\end{tabular}
}

\label{tab: trans_vs_rnn}
\end{centering}
\vspace{-4mm}
\end{table}

\vspace{-2mm}
\section{Conclusions and limitations}
In this paper, we present a novel multi-view depth estimation method for videos. There are two main contributions. First, we propose an Epipolar Spatio-Temporal transformer to associate temporal coherence with multiple estimated depth maps. Second, for better computational efficiency, we design a hybrid network for cost regularization, which disentangles the learning of 2D global context information from that of 3D local matching information. Experiments demonstrate
that our method outperforms the state-of-the-art in terms of depth accuracy and speed.

There are several limitations we plan to tackle in the future. First, since our method takes video frames with known camera poses as input, the accuracy of camera poses will influence the performance of our EST transformer. A general framework that jointly estimates camera pose and depth maps may alleviate this problem. Second, our work takes only a short video clip into consideration, rather than estimate depth maps in a global optimization scheme for further potential improvement of temporal coherence. 

\begin{acks}
We thank the anonymous reviewers for their valuable feedback. This work was supported by the General Research Fund (GRF) of the Research Grant Council of Hong Kong (17210718), ERC Consolidator Grant 4DReply (770784), and Lise Meitner Postdoctoral Fellowship
\end{acks}

{\small
\bibliographystyle{ieee_fullname}

\begin{thebibliography}{10}\itemsep=-1pt

\bibitem{bleyer2011patchmatch}
Michael Bleyer, Christoph Rhemann, and Carsten Rother.
\newblock Patchmatch stereo-stereo matching with slanted support windows.
\newblock In {\em Bmvc}, volume~11, pages 1--11, 2011.

\bibitem{chang2018pyramid}
Jia-Ren Chang and Yong-Sheng Chen.
\newblock Pyramid stereo matching network.
\newblock In {\em Proceedings of the IEEE Conference on Computer Vision and
  Pattern Recognition}, pages 5410--5418, 2018.

\bibitem{chen2018rethinking}
Richard Chen, Faisal Mahmood, Alan Yuille, and Nicholas~J Durr.
\newblock Rethinking monocular depth estimation with adversarial training.
\newblock {\em arXiv preprint arXiv:1808.07528}, 2018.

\bibitem{chen2016single}
Weifeng Chen, Zhao Fu, Dawei Yang, and Jia Deng.
\newblock Single-image depth perception in the wild.
\newblock In {\em Advances in neural information processing systems}, pages
  730--738, 2016.

\bibitem{cheng2019learning}
Xinjing Cheng, Peng Wang, and Ruigang Yang.
\newblock Learning depth with convolutional spatial propagation network.
\newblock {\em IEEE transactions on pattern analysis and machine intelligence},
  2019.

\bibitem{cs2018depthnet}
Arun CS~Kumar, Suchendra~M Bhandarkar, and Mukta Prasad.
\newblock Depthnet: A recurrent neural network architecture for monocular depth
  prediction.
\newblock In {\em Proceedings of the IEEE Conference on Computer Vision and
  Pattern Recognition Workshops}, pages 283--291, 2018.

\bibitem{dai2017scannet}
Angela Dai, Angel~X Chang, Manolis Savva, Maciej Halber, Thomas Funkhouser, and
  Matthias Nie{\ss}ner.
\newblock Scannet: Richly-annotated 3d reconstructions of indoor scenes.
\newblock In {\em Proceedings of the IEEE Conference on Computer Vision and
  Pattern Recognition}, pages 5828--5839, 2017.

\bibitem{eigen2015predicting}
David Eigen and Rob Fergus.
\newblock Predicting depth, surface normals and semantic labels with a common
  multi-scale convolutional architecture.
\newblock In {\em Proceedings of the IEEE international conference on computer
  vision}, pages 2650--2658, 2015.

\bibitem{eigen2014depth}
David Eigen, Christian Puhrsch, and Rob Fergus.
\newblock Depth map prediction from a single image using a multi-scale deep
  network.
\newblock In {\em Advances in neural information processing systems}, pages
  2366--2374, 2014.

\bibitem{fehn2004depth}
Christoph Fehn.
\newblock Depth-image-based rendering (dibr), compression, and transmission for
  a new approach on 3d-tv.
\newblock In {\em Stereoscopic Displays and Virtual Reality Systems XI}, volume
  5291, pages 93--104. International Society for Optics and Photonics, 2004.

\bibitem{fu2018deep}
Huan Fu, Mingming Gong, Chaohui Wang, Kayhan Batmanghelich, and Dacheng Tao.
\newblock Deep ordinal regression network for monocular depth estimation.
\newblock In {\em Proceedings of the IEEE Conference on Computer Vision and
  Pattern Recognition}, pages 2002--2011, 2018.

\bibitem{gallup2007real}
David Gallup, Jan-Michael Frahm, Philippos Mordohai, Qingxiong Yang, and Marc
  Pollefeys.
\newblock Real-time plane-sweeping stereo with multiple sweeping directions.
\newblock In {\em 2007 IEEE Conference on Computer Vision and Pattern
  Recognition}, pages 1--8. IEEE, 2007.

\bibitem{henry2014rgb}
Peter Henry, Michael Krainin, Evan Herbst, Xiaofeng Ren, and Dieter Fox.
\newblock Rgb-d mapping: Using depth cameras for dense 3d modeling of indoor
  environments.
\newblock In {\em Experimental robotics}, pages 477--491. Springer, 2014.

\bibitem{hirschmuller2007stereo}
Heiko Hirschmuller.
\newblock Stereo processing by semiglobal matching and mutual information.
\newblock {\em IEEE Transactions on pattern analysis and machine intelligence},
  30(2):328--341, 2007.

\bibitem{hosni2009local}
Asmaa Hosni, Michael Bleyer, Margrit Gelautz, and Christoph Rhemann.
\newblock Local stereo matching using geodesic support weights.
\newblock In {\em 2009 16th IEEE International Conference on Image Processing
  (ICIP)}, pages 2093--2096. IEEE, 2009.

\bibitem{hosni2012fast}
Asmaa Hosni, Christoph Rhemann, Michael Bleyer, Carsten Rother, and Margrit
  Gelautz.
\newblock Fast cost-volume filtering for visual correspondence and beyond.
\newblock {\em IEEE Transactions on Pattern Analysis and Machine Intelligence},
  35(2):504--511, 2012.

\bibitem{im2019dpsnet}
Sunghoon Im, Hae-Gon Jeon, Stephen Lin, and In~So Kweon.
\newblock Dpsnet: end-to-end deep plane sweep stereo.
\newblock {\em arXiv preprint arXiv:1905.00538}, 2019.

\bibitem{jones2008effects}
J~Adam Jones, J~Edward Swan, Gurjot Singh, Eric Kolstad, and Stephen~R Ellis.
\newblock The effects of virtual reality, augmented reality, and motion
  parallax on egocentric depth perception.
\newblock In {\em Proceedings of the 5th symposium on Applied perception in
  graphics and visualization}, pages 9--14, 2008.

\bibitem{kendall2017end}
Alex Kendall, Hayk Martirosyan, Saumitro Dasgupta, Peter Henry, Ryan Kennedy,
  Abraham Bachrach, and Adam Bry.
\newblock End-to-end learning of geometry and context for deep stereo
  regression.
\newblock In {\em Proceedings of the IEEE International Conference on Computer
  Vision}, pages 66--75, 2017.

\bibitem{kusupati2020normal}
Uday Kusupati, Shuo Cheng, Rui Chen, and Hao Su.
\newblock Normal assisted stereo depth estimation.
\newblock In {\em Proceedings of the IEEE/CVF Conference on Computer Vision and
  Pattern Recognition}, pages 2189--2199, 2020.

\bibitem{liu2019neural}
Chao Liu, Jinwei Gu, Kihwan Kim, Srinivasa~G Narasimhan, and Jan Kautz.
\newblock Neural rgb (r) d sensing: Depth and uncertainty from a video camera.
\newblock In {\em Proceedings of the IEEE Conference on Computer Vision and
  Pattern Recognition}, pages 10986--10995, 2019.

\bibitem{liu2015deep}
Fayao Liu, Chunhua Shen, and Guosheng Lin.
\newblock Deep convolutional neural fields for depth estimation from a single
  image.
\newblock In {\em Proceedings of the IEEE Conference on Computer Vision and
  Pattern Recognition}, pages 5162--5170, 2015.

\bibitem{long2020occlusion}
Xiaoxiao Long, Lingjie Liu, Christian Theobalt, and Wenping Wang.
\newblock Occlusion-aware depth estimation with adaptive normal constraints.
\newblock {\em arXiv preprint arXiv:2004.00845}, 2020.

\bibitem{luo2020consistent}
Xuan Luo, Jia-Bin Huang, Richard Szeliski, Kevin Matzen, and Johannes Kopf.
\newblock Consistent video depth estimation.
\newblock {\em arXiv preprint arXiv:2004.15021}, 2020.

\bibitem{ma2018modeling}
Jiaqi Ma, Zhe Zhao, Xinyang Yi, Jilin Chen, Lichan Hong, and Ed~H Chi.
\newblock Modeling task relationships in multi-task learning with multi-gate
  mixture-of-experts.
\newblock In {\em Proceedings of the 24th ACM SIGKDD International Conference
  on Knowledge Discovery \& Data Mining}, pages 1930--1939, 2018.

\bibitem{patil2020don}
Vaishakh Patil, Wouter Van~Gansbeke, Dengxin Dai, and Luc Van~Gool.
\newblock Don't forget the past: Recurrent depth estimation from monocular
  video.
\newblock {\em arXiv preprint arXiv:2001.02613}, 2020.

\bibitem{qi2018geonet}
Xiaojuan Qi, Renjie Liao, Zhengzhe Liu, Raquel Urtasun, and Jiaya Jia.
\newblock Geonet: Geometric neural network for joint depth and surface normal
  estimation.
\newblock In {\em Proceedings of the IEEE Conference on Computer Vision and
  Pattern Recognition}, pages 283--291, 2018.

\bibitem{schonberger2016structure}
Johannes~L Schonberger and Jan-Michael Frahm.
\newblock Structure-from-motion revisited.
\newblock In {\em Proceedings of the IEEE Conference on Computer Vision and
  Pattern Recognition}, pages 4104--4113, 2016.

\bibitem{shotton2013scene}
Jamie Shotton, Ben Glocker, Christopher Zach, Shahram Izadi, Antonio Criminisi,
  and Andrew Fitzgibbon.
\newblock Scene coordinate regression forests for camera relocalization in
  rgb-d images.
\newblock In {\em Proceedings of the IEEE Conference on Computer Vision and
  Pattern Recognition}, pages 2930--2937, 2013.

\bibitem{sinha2020depth}
Ayan Sinha, Zak Murez, James Bartolozzi, Vijay Badrinarayanan, and Andrew
  Rabinovich.
\newblock Depth estimation by learning triangulation and densification of
  sparse points for multi-view stereo.
\newblock {\em arXiv preprint arXiv:2003.08933}, 2020.

\bibitem{ummenhofer2017demon}
Benjamin Ummenhofer, Huizhong Zhou, Jonas Uhrig, Nikolaus Mayer, Eddy Ilg,
  Alexey Dosovitskiy, and Thomas Brox.
\newblock Demon: Depth and motion network for learning monocular stereo.
\newblock In {\em Proceedings of the IEEE Conference on Computer Vision and
  Pattern Recognition}, pages 5038--5047, 2017.

\bibitem{wang2018mvdepthnet}
Kaixuan Wang and Shaojie Shen.
\newblock Mvdepthnet: real-time multiview depth estimation neural network.
\newblock In {\em 2018 International Conference on 3D Vision (3DV)}, pages
  248--257. IEEE, 2018.

\bibitem{wang2015towards}
Peng Wang, Xiaohui Shen, Zhe Lin, Scott Cohen, Brian Price, and Alan~L Yuille.
\newblock Towards unified depth and semantic prediction from a single image.
\newblock In {\em Proceedings of the IEEE conference on computer vision and
  pattern recognition}, pages 2800--2809, 2015.

\bibitem{wang2019recurrent}
Rui Wang, Stephen~M Pizer, and Jan-Michael Frahm.
\newblock Recurrent neural network for (un-) supervised learning of monocular
  video visual odometry and depth.
\newblock In {\em Proceedings of the IEEE Conference on Computer Vision and
  Pattern Recognition}, pages 5555--5564, 2019.

\bibitem{wang2020deep}
Xin Wang, Fisher Yu, Lisa Dunlap, Yi-An Ma, Ruth Wang, Azalia Mirhoseini,
  Trevor Darrell, and Joseph~E Gonzalez.
\newblock Deep mixture of experts via shallow embedding.
\newblock In {\em Uncertainty in Artificial Intelligence}, pages 552--562.
  PMLR, 2020.

\bibitem{wang2019pseudo}
Yan Wang, Wei-Lun Chao, Divyansh Garg, Bharath Hariharan, Mark Campbell, and
  Kilian~Q Weinberger.
\newblock Pseudo-lidar from visual depth estimation: Bridging the gap in 3d
  object detection for autonomous driving.
\newblock In {\em Proceedings of the IEEE Conference on Computer Vision and
  Pattern Recognition}, pages 8445--8453, 2019.

\bibitem{xiao2013sun3d}
Jianxiong Xiao, Andrew Owens, and Antonio Torralba.
\newblock Sun3d: A database of big spaces reconstructed using sfm and object
  labels.
\newblock In {\em Proceedings of the IEEE International Conference on Computer
  Vision}, pages 1625--1632, 2013.

\bibitem{xu2019depth}
Yan Xu, Xinge Zhu, Jianping Shi, Guofeng Zhang, Hujun Bao, and Hongsheng Li.
\newblock Depth completion from sparse lidar data with depth-normal
  constraints.
\newblock In {\em Proceedings of the IEEE International Conference on Computer
  Vision}, pages 2811--2820, 2019.

\bibitem{yao2018mvsnet}
Yao Yao, Zixin Luo, Shiwei Li, Tian Fang, and Long Quan.
\newblock Mvsnet: Depth inference for unstructured multi-view stereo.
\newblock In {\em Proceedings of the European Conference on Computer Vision
  (ECCV)}, pages 767--783, 2018.

\bibitem{Yin2019enforcing}
Wei Yin, Yifan Liu, Chunhua Shen, and Youliang Yan.
\newblock Enforcing geometric constraints of virtual normal for depth
  prediction.
\newblock In {\em The IEEE International Conference on Computer Vision (ICCV)},
  2019.

\bibitem{zeng20163dmatch}
Andy Zeng, Shuran Song, Matthias Nie{\ss}ner, Matthew Fisher, Jianxiong Xiao,
  and Thomas Funkhouser.
\newblock 3dmatch: Learning local geometric descriptors from rgb-d
  reconstructions.
\newblock In {\em CVPR}, 2017.

\bibitem{zhang2009consistent}
Guofeng Zhang, Jiaya Jia, Tien-Tsin Wong, and Hujun Bao.
\newblock Consistent depth maps recovery from a video sequence.
\newblock {\em IEEE Transactions on pattern analysis and machine intelligence},
  31(6):974--988, 2009.

\bibitem{zhang2019exploiting}
Haokui Zhang, Chunhua Shen, Ying Li, Yuanzhouhan Cao, Yu Liu, and Youliang Yan.
\newblock Exploiting temporal consistency for real-time video depth estimation.
\newblock In {\em Proceedings of the IEEE International Conference on Computer
  Vision}, pages 1725--1734, 2019.

\bibitem{zhang2019learning}
Lianbo Zhang, Shaoli Huang, Wei Liu, and Dacheng Tao.
\newblock Learning a mixture of granularity-specific experts for fine-grained
  categorization.
\newblock In {\em Proceedings of the IEEE International Conference on Computer
  Vision}, pages 8331--8340, 2019.

\end{thebibliography}

}

\end{document}